\PassOptionsToPackage{table}{xcolor}
\documentclass[sigconf]{acmart}

\AtBeginDocument{%
  }

\copyrightyear{2024}
\acmYear{2024}
\setcopyright{rightsretained}
\acmConference[CIKM '24]{Proceedings of the 33rd ACM International Conference on Information and Knowledge Management}{October 21--25, 2024}{Boise, ID, USA}
\acmBooktitle{Proceedings of the 33rd ACM International Conference on Information and Knowledge Management (CIKM '24), October 21--25, 2024, Boise, ID, USA}
\acmDOI{10.1145/3627673.3679989}
\acmISBN{979-8-4007-0436-9/24/10}

\usepackage{amsfonts,bm}
\usepackage{cleveref}
\usepackage{enumitem}
\usepackage{mathtools}
\usepackage{xspace}

\newcommand{\name}{{STMAE}}
\definecolor{gray}{rgb}{0.85,0.85,0.85}

\makeatletter
\DeclareRobustCommand\onedot{\futurelet\@let@token\@onedot}
\def\@onedot{\ifx\@let@token.\else.\null\fi\xspace}

\def\ie{\emph{i.e}\onedot}

\def\etal{\emph{et al}\onedot}
\makeatother

\def\eqref#1{equation~\ref{#1}}

\def\1{\bm{1}}

\def\rmW{{\mathbf{W}}}

\def\mA{{\mathbf{A}}}

\def\mS{{\mathbf{S}}}

\def\mX{{\mathcal{X}}}
\def\mY{{\mathcal{Y}}}

\def\gE{{\mathcal{E}}}

\def\gG{{\mathcal{G}}}

\def\gL{{\mathcal{L}}}

\def\gR{{\mathcal{R}}}

\def\gV{{\mathcal{V}}}

\def\sR{{\mathbb{R}}}

\newcommand{\onenorm}[1]{\| #1 \|_1}

\begin{document}

\title{Revealing the Power of Masked Autoencoders in Traffic Forecasting}

\author{Jiarui Sun}
\authornote{Work done while at Visa Research. Our code is released at: \url{https://github.com/jsun57/STMAE}.}
\affiliation{%
  \institution{\mbox{University of Illinois Urbana-Champaign}}
  \city{Urbana}
  \state{IL}
  \country{USA}
}
\email{jsun57@illinois.edu}
\orcid{0000-0002-7113-081X}

\author{Yujie Fan}
\affiliation{%
  \institution{Visa Research}
  \city{Foster City}
  \state{CA}
  \country{USA}}
\email{yufan@visa.com}
\orcid{0000-0002-2635-9420}

\author{Chin-Chia Michael Yeh}
\affiliation{%
  \institution{Visa Research}
  \city{Foster City}
  \state{CA}
  \country{USA}}
\email{miyeh@visa.com}
\orcid{0000-0002-9807-2963}

\author{Wei Zhang}
\affiliation{%
  \institution{Visa Research}
  \city{Foster City}
  \state{CA}
  \country{USA}}
\email{wzhan@visa.com}
\orcid{0009-0001-7984-7241}

\author{Girish Chowdhary}
\affiliation{%
  \institution{\mbox{University of Illinois Urbana-Champaign}}
  \city{Urbana}
  \state{IL}
  \country{USA}
}
\email{girishc@illinois.edu}
\orcid{0000-0002-4657-307X}

\renewcommand{\shortauthors}{Sun et al.}

\begin{abstract}
    Traffic forecasting, crucial for urban planning, requires accurate predictions of spatial-temporal traffic patterns across urban areas. 
    Existing research mainly focuses on designing complex models that capture spatial-temporal dependencies among variables explicitly. 
    However, this field faces challenges related to data scarcity and model stability, which results in limited performance improvement.
    To address these issues, we propose Spatial-Temporal Masked AutoEncoders (STMAE), a plug-and-play framework designed to enhance existing spatial-temporal models on traffic prediction.
    STMAE consists of two learning stages.
    In the pretraining stage, an encoder processes partially visible traffic data produced by a dual-masking strategy, including biased random walk-based spatial masking and patch-based temporal masking. 
    Subsequently, two decoders aim to reconstruct the masked counterparts from both spatial and temporal perspectives. 
    The fine-tuning stage retains the pretrained encoder and integrates it with decoders from existing backbones to improve forecasting accuracy.
    Our results on traffic benchmarks show that STMAE can largely enhance the forecasting capabilities of various spatial-temporal models.
\end{abstract}



\maketitle

\section{Introduction}
Spatial-temporal forecasting is a pivotal task with applications in numerous fields, including human motion modeling \cite{DBLP:conf/iccv/MaoLSL19,DBLP:journals/corr/abs-2305-12554} and epidemic simulation \cite{DBLP:journals/corr/abs-2007-03113,DBLP:conf/cikm/DengWRWN20}. 
Within these applications, traffic forecasting emerges as a crucial area for urban planning and management. 
This particular aspect of spatial-temporal analysis focuses on predicting traffic flow and speed based on historical data, which is essential for optimizing transportation systems.
To make accurate predictions, state-of-the-art methods develop spatial-temporal models to capture complex interactions of traffic data over both space and time.
By jointly modeling spatial and temporal patterns, these models exhibit remarkable traffic prediction performance.

However, despite efforts, several issues impede the performance of spatial-temporal models.
First, \textit{data scarcity} in traffic datasets often leads to \textit{model overfitting}. 
Unlike domains such as computer vision which utilize massive datasets for training, traffic benchmarks typically cover only a few months and are limited to specific geographic locations. 
This constrained data scope makes it difficult for models to generalize well across varying traffic patterns, increasing the risk of overfitting.
Second, \textit{data incompleteness} poses challenges to \textit{model stability}. 
Traffic data, collected from sensors deployed across road networks, inevitably suffer from inconsistencies due to sensor malfunctions. 
These issues lead to gaps and errors in the data, resulting in unstable training conditions for spatial-temporal models, which ultimately compromises their performance.
\begin{figure}[!tb]
\centering
\includegraphics[width=1\linewidth]{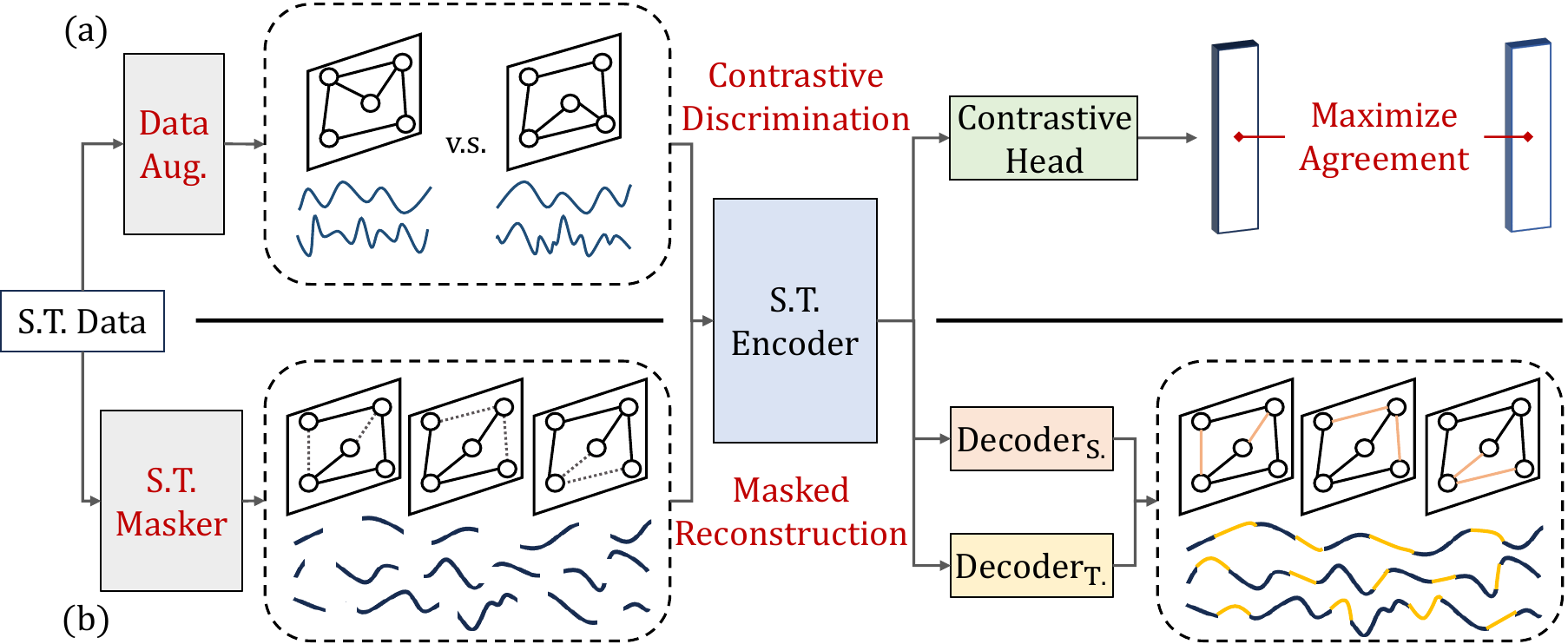}
\caption{ Illustration of SSL approaches: (a) Contrastive-based and (b) Mask-based frameworks for traffic forecasting.}
\Description{An illustration / comparison between contrastive-based and mask-based frameworks for traffic forecasting.}
\label{fig::overview}
\end{figure}

\begin{figure*}[!tb]
\centering
\includegraphics[width=0.9\linewidth]{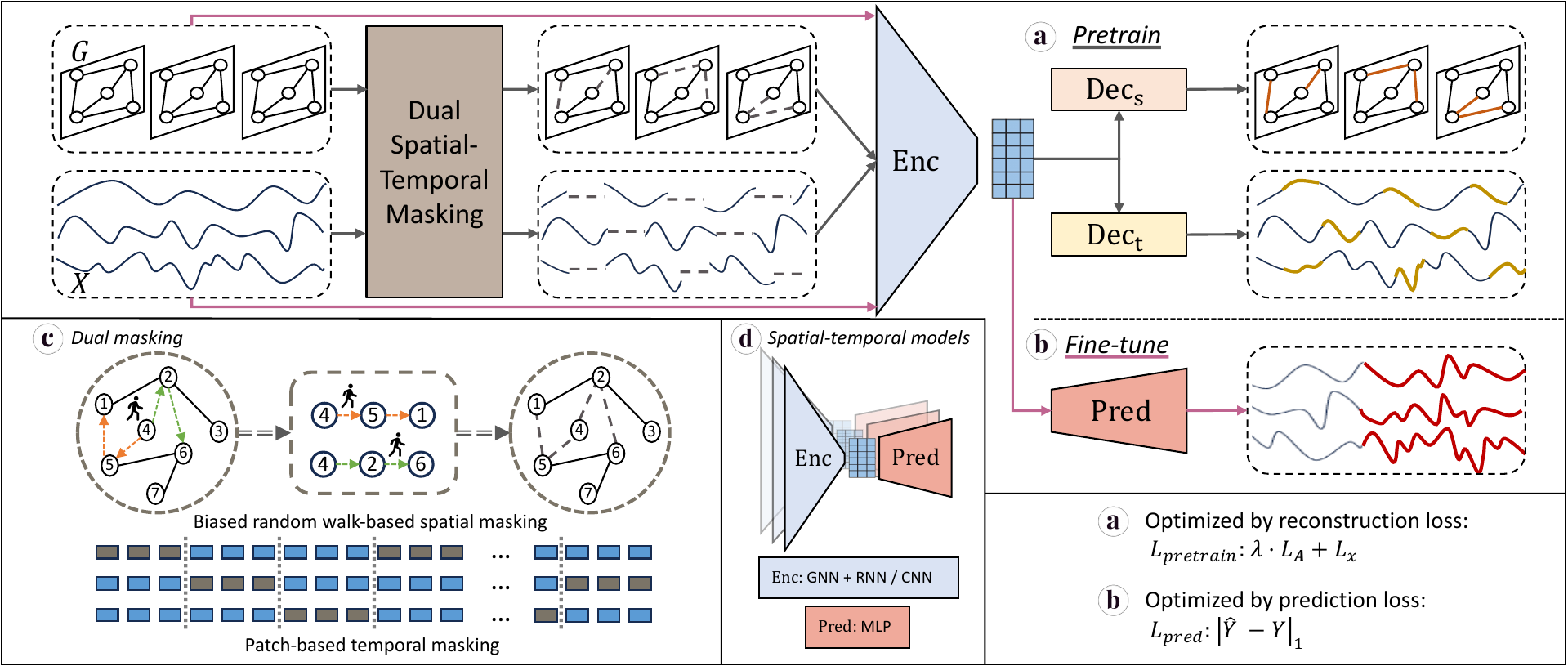}
\caption{The \name~framework, including the (a) pretraining and (b) fine-tuning stages. Specified by (c), We use a biased random walk-based spatial masking strategy on $\gG$, and a patch-based temporal masking strategy on $\mX$. After reconstruction, learning is guided jointly by $\gL_{\mA}$ and $\gL_{\mX}$. As shown in (d), \name~can be easily plugged into existing spatial-temporal models.}
\Description{A detailed illustration of the overall \name~framework.}
\label{fig::mdl}
\end{figure*}

Self-supervised learning (SSL) \cite{DBLP:conf/eccv/ZhangIE16} offers a promising avenue for mitigating the above challenges.
Contrastive SSL, for instance, shows potential by learning to contrast sample pairs, as depicted in \cref{fig::overview} (a). 
However, this method heavily depends on manually crafted heuristics to derive data samples, which are limited in the traffic scenario.
To overcome the limitation, we turn to generative SSL \cite{DBLP:journals/tkde/LiuZHMWZT23}, specifically Masked Autoencoders (MAE) \cite{DBLP:conf/cvpr/HeCXLDG22} as a more potent solution. 
Illustrated in \cref{fig::overview} (b), MAE focuses on reconstructing masked portions of input data to learn robust features, and can be seamlessly integrated to spatial-temporal models for traffic forecasting. 
Notably, this approach not only mitigates overfitting via self-reconstruction but also improves model stability by learning from incomplete data, thus enhancing prediction accuracy.

Inspired by MAE, we present \textit{Spatial-Temporal Masked AutoEncoders} (\name), a versatile framework that leverages the principles of MAE to elevate the capability of existing spatial-temporal models for traffic forecasting. 
\name, shown in \Cref{fig::mdl}, includes pretraining and fine-tuning stages.
Initially, an encoder from an existing spatial-temporal model, along with two decoders, reconstructs masked traffic data produced by a \textit{dual-masking strategy} that combines biased random walk-based spatial and patch-based temporal masking.
This challenges the encoder to acquire robust and predictive traffic features.
In the fine-tuning phase, the encoder is retained, and the decoders are replaced with the original predictor from the backbone for forecasting. 
\name~offers seamless integration to existing models without needing complex data augmentation, positioning itself as a flexible and powerful enhancement tool for traffic forecasting.
Our primary contributions are summarized below:

\begin{itemize}[leftmargin=*]
    \item We present \name, a versatile framework that seamlessly integrates with established spatial-temporal backbones, alleviating model overfitting and stability concerns. 
    \item We propose a dual-masking strategy, including biased random walk-based spatial masking and patch-based temporal masking.
    This establishes a challenging pretext task that encourages the encoder to acquire informative traffic data representations.
    \item We conducted extensive experiments on various traffic benchmarks, showing large performance enhancements over existing spatial-temporal models.
\end{itemize}

\section{Related Work}

\subsection{Spatial-Temporal Models}
\label{subsec::st}

By considering the interdependence among different variables, spatial-temporal models \cite{DBLP:conf/ijcai/WuPLJZ19,DBLP:journals/access/ZhangYL19a,DBLP:conf/kdd/FangLSX21,DBLP:conf/aaai/ZhengFW020,DBLP:conf/icml/LanMHWYL22} have shown remarkable traffic forecasting performance. 
For example,
Li \etal introduced DCRNN \cite{DBLP:conf/iclr/LiYS018}, which utilizes RNN-based graph diffusion convolution to capture spatial-temporal correlations.
Bai \etal proposed AGCRN \cite{DBLP:conf/nips/0001YL0020}, which incorporates several adaptive graph learning modules that dynamically capture sensor dependencies recurrently.
Wu \etal designed MTGNN \cite{DBLP:conf/kdd/WuPL0CZ20}, which learns spatial-temporal correlations among traffic sensors effectively using adaptive GNN and CNN-based modules.
Despite these innovations, the limited scale of traffic data often leads to model overfitting.
Addressing these challenges, Liu \etal introduced STGCL \cite{DBLP:conf/gis/0014LH0HZ22}, using contrastive SSL to enhance model learning from traffic network structures and temporal patterns. However, STGCL's reliance on specific data knowledge limits its application. In contrast, our STMAE framework provides a flexible, plug-and-play solution without needing such detailed prior information, offering a robust enhancement to existing models.

\subsection{Masked Autoencoders}
Masked autoencoders (MAE) are a generative SSL method for robust feature representation learning \cite{DBLP:conf/naacl/DevlinCLT19,DBLP:conf/cvpr/HeCXLDG22,DBLP:conf/nips/Feichtenhofer0L22}.
MAE employs an encoder to map masked inputs to latent representations and a decoder for reconstruction.
It has been adapted across domains: BERT \cite{DBLP:conf/naacl/DevlinCLT19} uses bidirectional context to model language through sentence masking, while the visual MAE \cite{DBLP:conf/cvpr/HeCXLDG22} is designed for image learning.
MAE has also been applied to graphs \cite{DBLP:conf/wsdm/TanL0CL0H23,DBLP:conf/aaai/0001DZZC23,DBLP:conf/sigir/YeX023,DBLP:conf/kdd/HouLCDYW022,DBLP:conf/kdd/LiWS0TZMZW23}. 
These methods underscore its potential as a unified representation learning method with minimal domain knowledge.
On a similar line of our work, STEP \cite{DBLP:conf/kdd/ShaoZWX22} uses a feature masking strategy with a Transformer model to encode temporal patterns from long-term traffic data.
Contrary to STEP with its specialized architecture, \name~is a more generalizable enhancer for various spatial-temporal backbones.

\section{Preliminaries}

\subsection{Traffic Forecasting}
\label{subsec::mtsfs}
Given traffic data $\mX \in \sR^{H\times N\times C}$ with $H$ frames of $N$ variables and $C$ features, traffic forecasting predicts future traffic $\hat{\mY} \in \sR^{F\times N\times C}$ for $F$ steps.
The variables in $\mX$ are spatially interrelated, represented by a graph $\gG = (\gV, \gE, \mA)$, with $\mA \in \sR^{N\times N}$ as the adjacency matrix representing connectivity.
The forecasting problem can be defined as $f_{\theta}(\mX, \mA) \rightarrow \hat{\mY}$,
where $f_{\theta}(\cdot)$ is the parameterized forecaster.

\subsection{Pipeline of Spatial-Temporal Models}
\label{subsec::pstm}

Existing spatial-temporal models often employ an encoder-decoder design. 
The \textit{encoder}, $\textrm{Enc}(\mX, \mA) \rightarrow \mS$, extracts complex spatial-temporal patterns from historical data into a hidden representation $\mS \in \sR^{N\times D}$, where $D$ is the hidden dimension. 
On the other hand, the \textit{predictor} (decoder), denoted as $\textrm{Pred}(\mS) \rightarrow \hat{\mY}$, focuses on making predictions based on the encoded state $\mS$.
In contrast to the encoder, $\textrm{Pred}(\cdot)$ is typically lightweight, often designed as a multilayer perceptron (MLP) with few layers.
Therefore, the pipeline of spatial-temporal models for traffic forecasting can be summarized as:
$f_{\theta}(\mX, \mA) \coloneqq {\textrm{Pred}}\bigl({\textrm{Enc}}(\mX, \mA)\bigr) \rightarrow \hat{\mY}$.
Lastly, the mean absolute error between predictions $\hat{\mY}$ and groundtruth $\mY$ is used as the objective to train spatial-temporal models.
This forecasting loss can be represented as $\gL_{\textrm{pred}} = \onenorm{\hat{\mY} - \mY}.$
\section{Methodology}


\name~implements a two-stage training process: pretraining and fine-tuning. 
During pretraining, \name~reconstructs masked traffic data using dual spatial-temporal masking. The fine-tuning stage utilizes the pretrained encoder to generate contextual representations for the predictor. \name~adopts existing spatial-temporal model components for both the encoder and predictor.

\begin{table*}[t]\small
\centering
\caption{Quantitative results of \name~compared with STGCL and various base models. Subscripts A, D, M correspond to the initials of the base models in which \name~and STGCL are coupled with. }
\begin{tabular}{c | r r r | r r r | r r r | r r r  }
\hline 
Datasets & \multicolumn{3}{c}{ PEMS03 } & \multicolumn{3}{c}{ PEMS04 } & \multicolumn{3}{c}{PEMS07} & \multicolumn{3}{c}{ PEMS08 }\\
\hline 
Method & MAE & MAPE & RMSE & MAE & MAPE & RMSE & MAE & MAPE & RMSE & MAE & MAPE & RMSE \\
\hline
\hline
AGCRN \cite{DBLP:conf/nips/0001YL0020} & 15.47 & 15.26 & \underline{27.06} & 19.39 & 13.24 & \textbf{31.07} & 20.64 & 8.80 & 34.19 & \underline{15.65} & \underline{10.33} & \underline{24.51} \\
$\mathrm{STGCL_A}$ & \underline{15.36} & \underline{14.85} & 27.15 & \underline{19.23} & \underline{13.01} & 31.36 & \underline{20.61} & \underline{8.74} & \underline{34.14} & 15.91 & 10.43 & 24.88 \\
\rowcolor{gray}
$\mathrm{STMAE_A}$ & \textbf{15.09} & \textbf{14.72} & \textbf{26.61} & \textbf{19.05} & \textbf{12.91} & \underline{31.32} & \textbf{20.13} & \textbf{8.53} & \textbf{33.79} & \textbf{15.01} & \textbf{9.79} & \textbf{23.97} \\
\hline
\hline
DCRNN \cite{DBLP:conf/iclr/LiYS018} & {15.76} & 15.69 & \underline{26.76} & 21.48 & 14.65 & 33.99 & {22.55} & {9.78} & {35.24} & 16.63 & 10.78 & {26.01} \\
footnotesize$\mathrm{STGCL_D}$ & \textbf{15.64} & \underline{15.68} & 27.08 & \underline{21.23} & \underline{14.57} & \underline{33.60} & \underline{22.34} &  \underline{9.68} & \underline{35.21} & \underline{16.51} & \underline{10.70} & \underline{25.93} \\
\rowcolor{gray}
$\mathrm{STMAE_D}$ & \underline{15.74} & \textbf{15.44} & \textbf{26.73} & \textbf{21.20} & \textbf{14.23} & \textbf{33.57} & \textbf{22.12} & \textbf{9.45} & \textbf{34.98} & \textbf{16.36} & \textbf{10.68} & \textbf{25.76} \\
\hline
\hline
MTGNN \cite{DBLP:conf/kdd/WuPL0CZ20} & 14.94 & 16.02 & \underline{25.29} & 19.02 & \underline{13.32} & 30.95 & 20.83 & 9.00 & \underline{33.77} & 15.44 & 10.35 & \underline{24.30} \\
$\mathrm{STGCL_M}$ & \underline{14.87} & \underline{15.37} & 25.53 & \underline{18.94} & 13.34 & \underline{30.79} & \underline{20.72} & \underline{8.95} & 33.78 & \underline{15.39} & \underline{10.13} & 24.32 \\
\rowcolor{gray}
$\mathrm{STMAE_M}$ & \textbf{14.84} & \textbf{14.15} & \textbf{24.95} & \textbf{18.87} & \textbf{12.78} & \textbf{30.28} & \textbf{20.57} & \textbf{8.90} & \textbf{33.47} & \textbf{15.03} & \textbf{9.82} & \textbf{24.08} \\
\hline
\hline
\end{tabular}
\label{tab::quan}
\end{table*}

\subsection{Pretraining}
\label{subsec::pt}
The pretraining stage uncovers spatial-temporal patterns in traffic data via self-supervised masking. 
First, we enhance existing encoders $\textrm{Enc}(\cdot)$ with a dual masking module.
Given the spatial-temporal interactions in traffic data, we combine spatial and temporal masking rather than isolating one domain. For the masking approach, spatially, we use biased random walk-based masking instead of simple uniform masking to preserve and challenge structural integrity of traffic networks. Temporally, we employ patch-based masking, dividing data into non-overlapping patches and masking these, which proves more challenging than masking single timesteps. Each component is detailed below.

\paragraph{Spatial Masking.}
While existing graph MAEs often uniformly mask a subset of relations, we enhance the challenge by using paths as the basic unit of masking to capture more complex structures. 
Leveraging a \textit{biased random walk-based spatial masking} strategy inspired by \cite{DBLP:conf/kdd/GroverL16}, we generate paths using a biased random walker that blends breadth-first and depth-first sampling, enabling deeper graph exploration. 
Relations for masking $\gE_{\textrm{mask}}$ with size $|\gE| \cdot p_s$ are selected from $\gE$ as:
\begin{equation}
\gE_{\textrm{mask}} \sim \textrm{BiasedRandomWalk}(\gE, \gR, p, q),
\end{equation}
where $p_s$ is the spatial masking ratio, $\gR \in \gV$ is the set of root nodes, and $p$ and $q$ are hyperparameters controlling the walk dynamics. Masked relations are then set to zero in the adjacency matrix $\mA$:
\begin{equation}
\mA_{uv} \leftarrow \begin{cases} 0 & (u, v) \in \gE_{\textrm{mask}}, \\ \mA_{uv} & \textrm{otherwise.}\end{cases}
\end{equation}
Our spatial masking strategy enhances the reconstruction task's complexity and the encoder’s capacity to infer hidden structure.

\paragraph{Temporal Masking.} 

With relatively low information density, traffic data can be easily recovered by interpolation, especially when temporal masking areas are sparse.
To counteract this, we implement \textit{patch-based temporal masking}. 
We partition the original data $\mX$ into $P$ nonoverlapping temporal patches, each of length $L$, and selectively mask a subset of these patches. 
Masked patches are then substituted with a shared, learnable mask token. 
Let $\{\mX_i\}_{i=1}^{P}$ represent the embedding patches with $\mX_i \in \sR^{L\times N\times D}$. Temporal masking is defined as:
\begin{equation}
 \mX_i \leftarrow \begin{cases} \mathcal{M} & r \sim \textrm{Bernoulli}(p_t), \ r = 1, \\ \mX_i, & \textrm{otherwise,}\end{cases}
\end{equation}
where $\mathcal{M}$ is the mask token and $p_t$ is the temporal masking ratio.

\paragraph{Decoders.} 
Second, two lightweight decoders are used for reconstruction.
The decoders' goal is to reconstruct both $\mA$ and $\mX$ simultaneously.
After masking, the masked inputs $\mA_{M}$ and $\mX_{M}$ are encoded by $\textrm{Enc}(\cdot)$ into a state $\mS$. 
Now, instead of predicting $\hat{\mY}$ directly, the goal is to reconstruct the original inputs from both spatial and temporal dimensions using two specialized decoders: $\textrm{Dec}_t(\cdot)$ for data and $\textrm{Dec}_s(\cdot)$ for structural dependencies:
\begin{equation}
    \mS = {\textrm{Enc}}(\mX_M, \mA_M), \hat{\mX} = {\textrm{Dec}_t}(\mS), \hat{\mA} = {\textrm{Dec}_s}(\mS), 
\end{equation}
where $\hat{\mX}$ and $\hat{\mA}$ are the reconstructed data and adjacency matrix. 
The spatial decoder, $\textrm{Dec}s(\cdot)$, performs a linear transformation followed by a sigmoid operation:
\begin{equation}
\hat{\mA}{uv} = \textrm{Sigmoid}\bigl((\mS \rmW)(\mS \rmW)^{T}{uv}\bigr),
\end{equation}
with $\rmW \in \sR^{D \times D}$ as a trainable matrix. The temporal decoder, $\textrm{Dec}_t(\cdot)$, uses a linear layer to map $\mS$ back to its original dimensions. This asymmetric encoder-decoder setup is designed to maximize the encoder's ability to capture detailed spatial-temporal interactions.


\paragraph{Reconstruction Targets.} 
To guide spatial-temporal reconstruction, we leverage two objectives, consisting of a classification loss $\gL_{\mA}$ and a regression loss $\gL_{\mX}$ tailored to adjacency matrix $\mA$ and the data $\mX$, respectively.
Specifically, $\gL_{\mA}$ aims to reconstruct the masked walks using a classification objective.  
Formally, we have:
\begin{equation}
    \gL_{\mA} = - \frac{1}{\left|\gE_{\textrm{mask}}\right|} \sum_{(u, v) \in \gE_{\textrm{mask}}} \log \hat{\mA}_{uv}.
\end{equation}
On the other hand, $\gL_{\mX}$ computes the mean absolute error of the masked patches between $\mX$ and the reconstruction $\hat{\mX}$:
\begin{equation}
    \gL_{\mX} = \onenorm{\hat{\mX}_{[M]} - \mX_{[M]}},
\end{equation}
where the subscript $[M]$ denotes the masked area index. In line with recent work \cite{DBLP:journals/corr/abs-2201-02534}, we only compute losses over the masked portions.
The overall objective of the pretraining stage is:
\begin{equation}
    \gL_{\textrm{pretrain}} = \lambda \cdot \gL_{\mA} + \gL_{\mX},
\end{equation}
where $\lambda$ is a non-negative hyperparameter trading off the spatial reconstruction loss and the temporal reconstruction loss.

\subsection{Fine-tuning}
\label{subsec::ft}

After pretraining the encoder $\textrm{Enc}(\cdot)$ with the reconstruction objectives, we fine-tune $\textrm{Enc}(\cdot)$ with the original predictor $\textrm{Pred}(\cdot)$ obtained from spatial-temporal models to predict $\hat{\mY}$.
In this stage, we provide $\textrm{Enc}(\cdot)$ with the complete spatial-temporal data without masking, as opposed to the pretraining stage and discard the spatial decoder $\textrm{Dec}_s(\cdot)$ and the temporal decoder $\textrm{Dec}_t(\cdot)$.
This fine-tuning process aligns with the training pipeline described in \cref{subsec::pstm}, aiming to optimize the forecasting loss $\gL_{\textrm{pred}}$.

\section{Experiments}

\subsection{Experimental Setup}
\label{subsec::setup}

\paragraph{Datasets.}
Four traffic benchmarks—PEMS03, PEMS04, PEMS07, and PEMS08 \cite{DBLP:conf/aaai/SongLGW20,DBLP:conf/gis/0014LH0HZ22,DBLP:conf/icml/LanMHWYL22}—are used to assess \name's effectiveness. 
Each dataset is split into training, validation, and test sets in a 6:2:2 ratio chronologically. 
The training set supports both pretraining and fine-tuning phases, consistent with standard practices \cite{zeng2021contrastive}.  
For a fair comparison with previous methods, we predict the next 12 steps based on 12-step historical traffic flow data.


\paragraph{Base Models and Metrics.}
\name~is evaluated against STGCL \cite{DBLP:conf/gis/0014LH0HZ22}, which uses contrastive SSL to enhance spatial-temporal models on traffic forecasting. 
Within \name, we incorporate widely recognized models DCRNN \cite{DBLP:conf/iclr/LiYS018}, AGCRN \cite{DBLP:conf/nips/0001YL0020}, and MTGNN \cite{DBLP:conf/kdd/WuPL0CZ20} to assess their performance improvements with our approach. 
DCRNN requires a predefined graph as input, whereas AGCRN and MTGNN dynamically learn the graph structure during training.
We adopt three metrics, including mean absolute error (MAE), root mean squared error (RMSE), and mean absolute percentage error (MAPE).

\paragraph{Implementation Details.}
\name~is a versatile framework that can be seamlessly integrated into existing spatial-temporal backbones, in which the encoder $\textrm{Enc}(\cdot)$ and the predictor $\textrm{Pred}(\cdot)$ are taken directly from the base models.
Therefore, our framework avoids extensive hyper-parameter tuning, such as adjusting model depth, batch size, hidden dimension and learning rate, which are commonly practiced to obtain optimal results for spatial-temporal models.
For hyper-parameters required for \name, we conduct a grid search to determine few hyper-parameters: $L$ from $\{2, 3\}$, $p$, $q$, $\lambda$ from $\{0.5, 1, 2, 4\}$ and the masking ratios from $20\%$ to $80\%$ based on the validation MAE performance.
For all datasets, we pretrain \name~for 100 epochs, followed by 100 epochs of fine-tuning.


\subsection{Results}
\label{subsec::result}

\paragraph{Quantitative Results.}

\Cref{tab::quan} summarizes the experimental results averaged across all evaluation timesteps on four datasets.
Through analysing the results, we have the following observations:
(1) Demonstrated by the comparison among three base models, AGCRN and MTGNN, that dynamically learn the graph from the data, tend to outperform DCRNN, which relies on a predefined graph. 
(2) SSL can enhance the capability of spatial-temporal models for the traffic forecasting task, as both STGCL and \name~achieve better performance than the backbones at most scenarios.
(3) Our proposed \name, which leverages generative-based SSL through masking reconstruction, always outperforms STGCL, which is contrastive learning-based.
More importantly, it does not require complex data augmentation processes that STGCL heavily relies on.
This shows the efficacy of our framework.

\begin{figure}[!tb]
\centering
\includegraphics[width=1.0\linewidth]{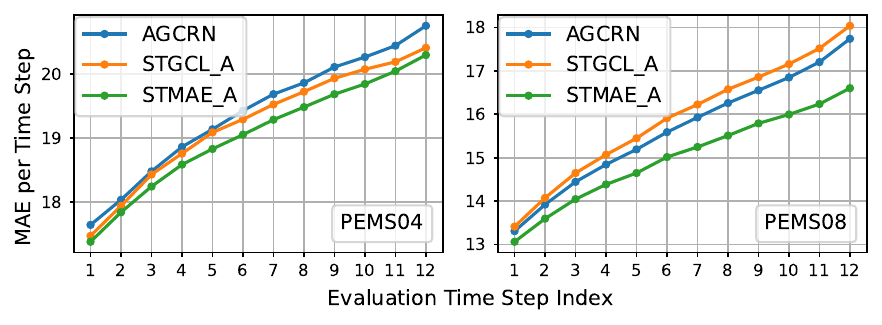}
\caption{Per-step MAE results of $\textrm{\name}_{\textrm{A}}$ compared with its corresponding base model AGCRN and $\textrm{STGCL}_{\textrm{A}}$.}
\Description{}
\label{fig::per_step}
\end{figure}
%

To gain a deeper insight into the predictive capabilities of \name, we illustrate the per-step MAE results of $\textrm{\name}_{\textrm{A}}$ and its corresponding base model and $\textrm{STGCL}_{\textrm{A}}$ in \cref{fig::per_step}.
Upon analysing the results, we have the following conclusions:
(1) The prediction error is positively correlated with the forecasting step as MAE continues to increase as the forecasting step becomes larger.
This indicates the performance degradation issue of traffic forecasting.
(2) $\textrm{\name}_{\textrm{A}}$ consistently outperforms $\textrm{STGCL}_{\textrm{A}}$ and their respective base models in terms of MAE at each timestep.
(3) Importantly, as evaluation forecasting timestep increases, the performance gap between \name~and the two variants gradually widens.
This suggests that \name, aided by masked self-supervision, can effectively alleviates the performance degradation issue.

\begin{figure}[!tb]
\centering
\includegraphics[width=1.0\linewidth]{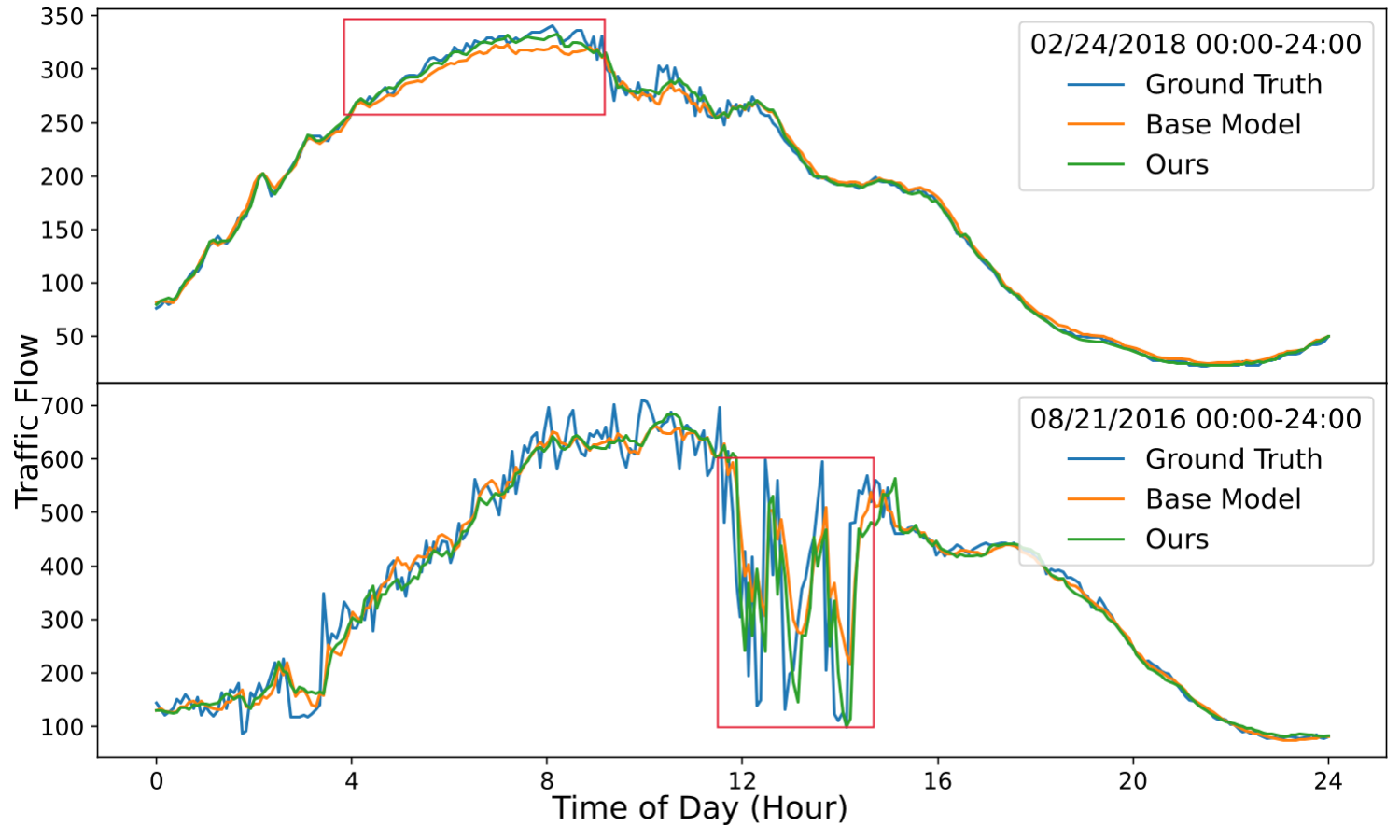}
\caption{Visualization of one-hour-ahead predictions on
two snapshots from PEMS04 and PEMS08 test sets.}
\Description{}
\label{fig::vis}
\end{figure}

\paragraph{Qualitative Visualization.} 
To study \name's predictive ability qualitatively, we randomly select two sensors from PEMS04 and PEMS08, and visualize one-hour-ahead predictions of $\textrm{\name}_{\textrm{A}}$ and AGCRN against the groundtruth on a test snapshot in \cref{fig::vis}.
We observe that \name~outperforms the counterpart base model both in areas with gradual or sudden changes (red rectangles), showing \name's efficacy in adapting to various spatial-temporal patterns. 

\subsection{Ablation Study}
\label{subsec::ablation}

In \name, a dual-masking strategy is proposed for model pretraining, targeting both spatial and temporal dimensions. To assess this, we introduced three variants: $\textrm{STMAE}_{\textrm{NT}}$ (no temporal masking), $\textrm{STMAE}_{\textrm{NS}}$ (no spatial masking), and $\textrm{STMAE}_{\textrm{U}}$ (uniform spatial-temporal masking).
All ablation results are presented in \cref{tab:ablation}, with AGCRN serving as the backbone and PEMS04, PEMS08 as evaluation datasets. 
The experimental results yield the following insights:
(1) Both $\textrm{STMAE}_{\textrm{NT}}$ and $\textrm{STMAE}_{\textrm{NS}}$ improve MAE performance, confirming the efficacy of individual spatial and temporal masking in the pretraining phase.
(2) $\textrm{STMAE}_{\textrm{U}}$ outperforms $\textrm{STMAE}_{\textrm{NT}}$ and $\textrm{STMAE}_{\textrm{NS}}$ in terms of MAE. 
This suggests that jointly applying masking techniques from both spatial and temporal perspectives is more advantageous for accurate prediction.
(3) \name~achieves superior performance compared to the base model and all its variants in terms of MAE, and it consistently ranks either first or second in terms of MAPE and RMSE metrics. 
This underscores the advantages of our approach, where the dual masking strategy can create challenging pretext tasks for spatial-temporal models, thus improving their traffic forecasting capacities.

\begin{table}[t]\footnotesize
\centering
\caption{Ablations of $\textrm{STMAE}_{\textrm{A}}$ on PEMS04 and PEMS08.} 
    \begin{tabular}{  c | ccc | ccc } 
      \hline
       & \multicolumn{3}{c}{ PEMS04 } & \multicolumn{3}{c}{ PEMS08 }\\
      \hline
      Variant  & MAE & MAPE & RMSE & MAE & MAPE & RMSE \\
      \hline
      Base Model & 19.39 & 13.24 & \textbf{31.07} & 15.65 & 10.33 & 24.51\\
      $\textrm{\name}_{\textrm{NT}}$ & 19.27 & 13.32 & \textbf{31.07} & 15.41 & 10.08 & 24.26\\
      $\textrm{\name}_{\textrm{NS}}$ & 19.27& \textbf{12.85} & 31.34 & 15.26 & \underline{9.89} & 24.43 \\
      $\textrm{\name}_{\textrm{U}}$ & \underline{19.11} & 12.96 & 31.67 & \underline{15.09} & 10.03 & \underline{24.22}\\
      \hline 
      \rowcolor{gray}
      \name & \textbf{19.05} & \underline{12.91} & \underline{31.32} & \textbf{15.01} & \textbf{9.79} & \textbf{23.97} \\
      \hline
    \end{tabular}
\label{tab:ablation}
\end{table}

\begin{figure}[!tb]
\centering
\includegraphics[width=1.0\linewidth]{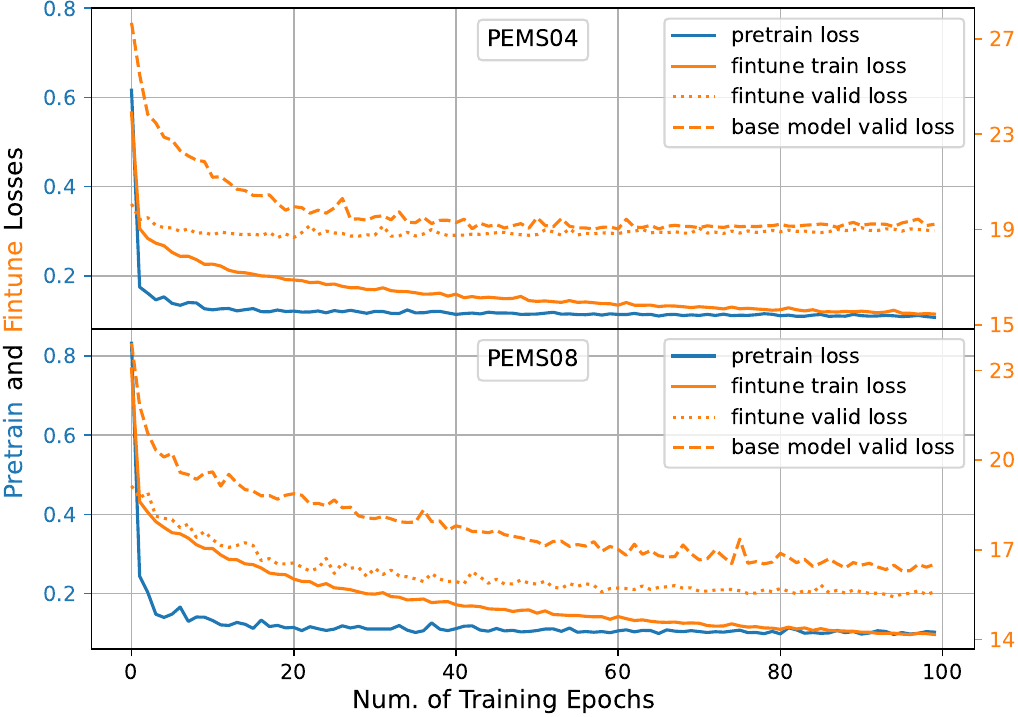}
\caption{Training and validation processes of $\textrm{\name}_{\textrm{A}}$ and AGCRN on PEMS04 and PEMS08. {Both pretraining and fine-tuning are performed for 100 epochs.}}
\Description{}
\label{fig::losses}
\end{figure}

\subsection{Stability Study}
\label{subsec::stab}
This section investigates \name's stability with an in-depth analysis of its learning behavior on PEMS04 and PEMS08 datasets. 
In \Cref{fig::losses}, we plot the learning curves for both the pretraining and fine-tuning stages of $\textrm{\name}_{\textrm{A}}$, alongside the learning curve for the AGCRN backbone.
Our analysis of the plot leads us to the following conclusions:
(1) Both the pretraining and fine-tuning stages of \name~are stable, as both training curves of \name~(\ie, pretrain loss and fine-tune train loss) depict a smoothly decreasing trend.
(2) The pretraining stage proves advantageous for \name. 
{Specifically, for the PEMS04 dataset, \name~exhibits a significantly lower initial validation loss compared to the base model, indicating a more effective starting point for fine-tuning.
For the PEMS08 dataset, \name~not only starts with a lower initial validation loss but also maintains this lead to achieve a substantially lower final validation loss. This underscores the effectiveness of the dual reconstruction objective employed during the pretraining phase, which enables \name~to initialize more effectively.}
(3) In contrast to \name, the validation curves of the base model exhibit a more erratic fluctuation behavior with a higher MAE level. 
In summary, all three observations above signify the stability of \name.

\subsection{Masking Ratio Sensitivity}
\label{subsec::sensitivity}

\begin{figure}[!tb]
\centering
\includegraphics[width=1.0\linewidth]{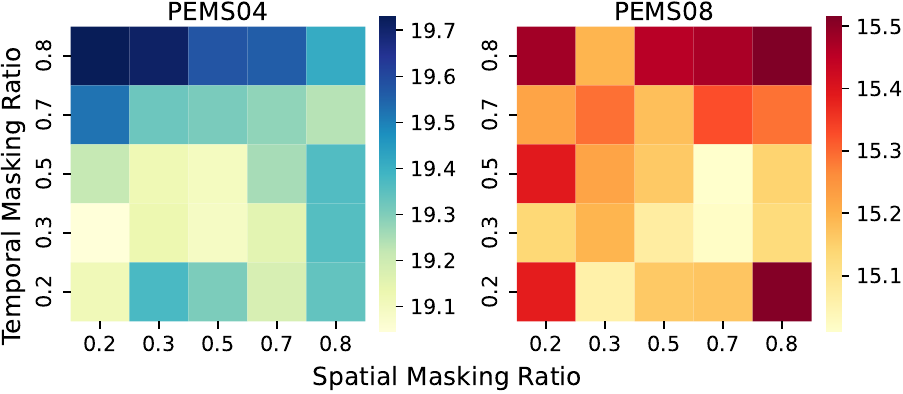}
\caption{Masking ratio sensitivity analysis of $\textrm{\name}_{\textrm{A}}$ on PEMS04 and PEMS08. Heatmaps show test MAE scores.}
\label{fig::sensitivity}
\Description{}
\end{figure}

In this section, we examine \name’s sensitivity to its key hyper-parameters, \ie, the spatial and temporal masking ratios.
We explore the impact of varying these ratios from 20\% to 80\% when coupling STMAE with AGCRN. 
Test MAE results for PEMS04 and PEMS08 datasets are shown in \cref{fig::sensitivity}. 
The heatmaps indicate that \name~achieves the best performance with a temporal masking ratio of 30\% for both datasets. 
Spatially, it peaks at a 70\% masking ratio for PEMS08 and 30\% for PEMS04. 
We hypothesis that this difference is attributed to the greater structural density of PEMS08's traffic network (0.01 for PEMS08 vs.\ 0.004 for PEMS04).
Notably, \name~underperforms when masking ratios are either too high or too low.
This observation, in conjunction with our ablation study, underscores the efficacy of our dual masking strategy in enhancing spatial-temporal model performance in traffic forecasting.
\section{Conclusion}
Spatial-temporal models are effective for traffic forecasting but often face challenges related to overfitting and instability due to data scarcity and incompleteness. 
In response, we explore generative SSL and propose a novel framework \name~based on masking reconstruction. 
\name~integrates seamlessly into existing traffic forecasting spatial-temporal frameworks, structured around pretraining and fine-tuning phases. 
During pretraining, our novel dual masking strategy—utilizing biased random walk-based spatial and patch-based temporal masking—presents rigorous challenges that prompt the encoder to learn robust spatial-temporal patterns. 
In the fine-tuning stage, we leverage the pretrained encoder with the original predictor to improve forecast accuracy. 
Comprehensive evaluations on multiple traffic benchmarks affirm that \name~greatly enhances the performance of traffic forecasting models. 
Ablation studies further validate the effectiveness of our dual-masking strategy in addressing the specific challenges of traffic prediction.

\bibliographystyle{ACM-Reference-Format}
\bibliography{sample-base}

\clearpage
\appendix

\section{Dataset Details}
\label{appendix:dataset}

Four widely used traffic datasets\footnote{\url{https://github.com/Davidham3/STSGCN/}} are selected to evaluate the effectiveness of \name~on the traffic forecasting task.
These include:
\begin{itemize}
    \item PEMS03 \cite{DBLP:conf/aaai/SongLGW20}: The dataset contains traffic flow data in the Bay Area. 
    307 sensors are used for data collection starting from 09/01/2018 to 12/01/2018.
    \item PEMS04 \cite{DBLP:conf/aaai/GuoLFSW19}: The dataset contains traffic flow data in the Bay Area. 
    There are 307 sensors and the period of data ranges from 01/01/2018 to 02/28/2018.
    \item PEMS07 \cite{DBLP:conf/aaai/SongLGW20}: The dataset contains traffic flow information from California.
    The data is collected by 883 sensors from 05/01/2017 to 08/07/2017.
    \item PEMS08 \cite{DBLP:conf/aaai/GuoLFSW19}: The dataset contains traffic flow information collected from 170 sensors in the San Bernardino area from 07/01/2016 to 08/31/2016.
\end{itemize}

The number of variables (sensors) $N$, number of edges, the density of the predefined graph $\gG$ and the time range are summarized in \cref{tab:dataset_stat}.
We follow the practices in \cite{DBLP:conf/gis/0014LH0HZ22} to construct the predefined graphs that are necessary for DCRNN, that the adjacency matrix $\mA$ is constructed by road-network distance using Gaussian thresholding \cite{DBLP:journals/spm/ShumanNFOV13}.
Formally, 
\begin{equation}
    \mA_{uv}=\begin{cases}\exp \left(-\frac{\textrm{dist}\left(u, v\right)^2}{\sigma^2}\right), & \textrm{if } \textrm{dist}\left(u, v\right) \leqslant k \\ 0, & \textrm{otherwise},
    \end{cases}
\end{equation}
where $\textrm{dist}(u, v)$ denotes the road network distance between sensor $u$ and $v$, $\sigma$ denotes distance standard deviation, and $k=0.1$ is the threshold.
The training, validation, and testing dataset partition split is performed with a ratio of 6:2:2 for all four datasets, and the partition details are provided in \cref{tab:data_pt}, where each instance represents a 5-minute traffic recording.

\begin{table}[htbp]\small
\centering
\caption{Dataset statistics.}
\begin{tabular}{c|cccc}
\hline Datasets & \#Sensors & \#Edges & Density & Time Range \\
\hline
PEMS03 & 358 & 442 & 0.007 & 09/18 - 11/18\\
PEMS04 & 307 & 209 & 0.004 & 01/18 - 02/18\\
PEMS07 & 883 & 790 & 0.002 & 05/17 - 08/17\\
PEMS08 & 170 & 137 & 0.01 & 07/16 - 08/16\\
\hline
\end{tabular}
\label{tab:dataset_stat}
\end{table}

\begin{table}[htbp]\small
\centering
\caption{Dataset partition details.}
\begin{tabular}{c|cccc}
\hline Datasets & \#Training & \#Validation & \#Testing & \#Instances  \\
\hline
PEMS03 & 15,724 & 5,242 & 5,242 & 26,208 \\
PEMS04 & 10,195 & 3,398 & 3,399 & 16,992 \\
PEMS07 & 16,934 & 5,645 & 5,645 & 28,224  \\
PEMS08 & 10,713 & 3,571 & 3,572 & 17,856 \\
\hline
\end{tabular}
\label{tab:data_pt}
\end{table}

\section{Base Model Details}
\label{appendix:baseline}

To study \name's effectiveness in enhancing various backbones for traffic forecasting, three widely recognized spatial-temporal base models are used. 
These include:
\begin{itemize}
    \item AGCRN \cite{DBLP:conf/nips/0001YL0020}: a pioneer traffic forecasting framework which avoids relying on pre-defined graphs by learning the dependency graph dynamically from data.
    From the spatial side, AGCRN learns an adaptive graph on which graph convolution operates. 
    Temporally, this adaptive graph convolution module is embedded in RNN to generate the state embeddings which are used for forecasting.
    \item DCRNN \cite{DBLP:conf/iclr/LiYS018}: an early traffic forecasting framework incorporating the graph concept for spatial modeling. 
    After graph construction, DCRNN learns spatial dependencies via bidirectional random walks on the graph, and the temporal correlations are modeled through RNNs with scheduled sampling.
    As a pioneer work, DCRNN is not as performative when compared to the other two base models.
    \item MTGNN \cite{DBLP:conf/kdd/WuPL0CZ20}: a lightweight spatial-temporal forecasting framework. 
    Similar to AGCRN, MTGNN also extracts relations among variables adaptively through a graph learning module. 
    Differently, MTGNN relies on interleaved graph convolution and temporal convolution modules to capture the spatial-temporal dependencies within the data. 
    Without utilizing recurrent structures, MTGNN is both more lightweight and computationally efficient compared with AGCRN and DCRNN.
\end{itemize}

The hyper-parameters of all three base models are used from their respective papers. 
For AGCRN, a two-layer RNN is used for encoding and one linear layer is used for prediction. 
The hidden dimension is $64$, and the learning rate is $0.003$.
For DCRNN, both the encoder and decoder consist of two-layer RNNs with hidden dimension $64$.
With an initial learning rate of $0.01$, scheduled sampling is employed with decay probability $\epsilon_i=\frac{\mu}{\mu+\exp (i / \mu)}$ at the $i^{th}$ training step, where $\mu=2000$ is the convergence rate.
For MTGNN, three graph convolution modules and three temporal convolution modules are interleaved to learn spatial-temporal correlations with node embedding dimension of $40$.
The learning rate and gradient clipping is set to $0.001$ and $5$, respectively.

Besides \name, these base models are also integrated with STGCL \cite{DBLP:conf/gis/0014LH0HZ22}, a contrastive learning-based spatial-temporal model enhancer.
STGCL defines an auxiliary contrastive objective, which is jointly optimized alongside the main forecasting loss.
This contrastive objective is calculated based on positive and negative samples, which are generated based on various data augmentation techniques guided by predefined heuristics.
Following the original paper setup, we select input masking from \{edge masking, input masking, temporal shifting, input smoothing\} as base augmentation technique with ratio of $0.01$ and a negative filtering threshold of $60$ minutes. 
Based on paper recommendations, STGCL is trained with a two-layer multilayer perceptron with batch normalization layer as the contrastive projection head.

\section{Per-Step Results}
\label{appendix:perstep}
\begin{figure}[htbp]
\centering
\includegraphics[width=1.0\linewidth]{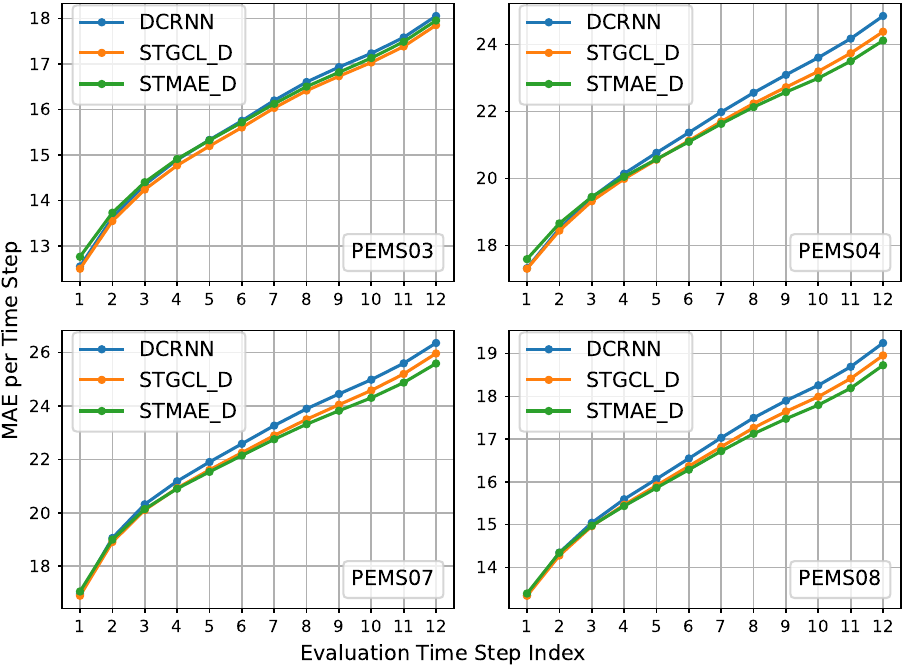}
\caption{Per-step MAE results of $\textrm{\name}_{\textrm{D}}$ compared with its corresponding base model DCRNN and $\textrm{STGCL}_{\textrm{D}}$.}
\label{fig::per_step_d}
\Description{}
\end{figure}
\begin{figure}[htbp]
\centering
\includegraphics[width=1.0\linewidth]{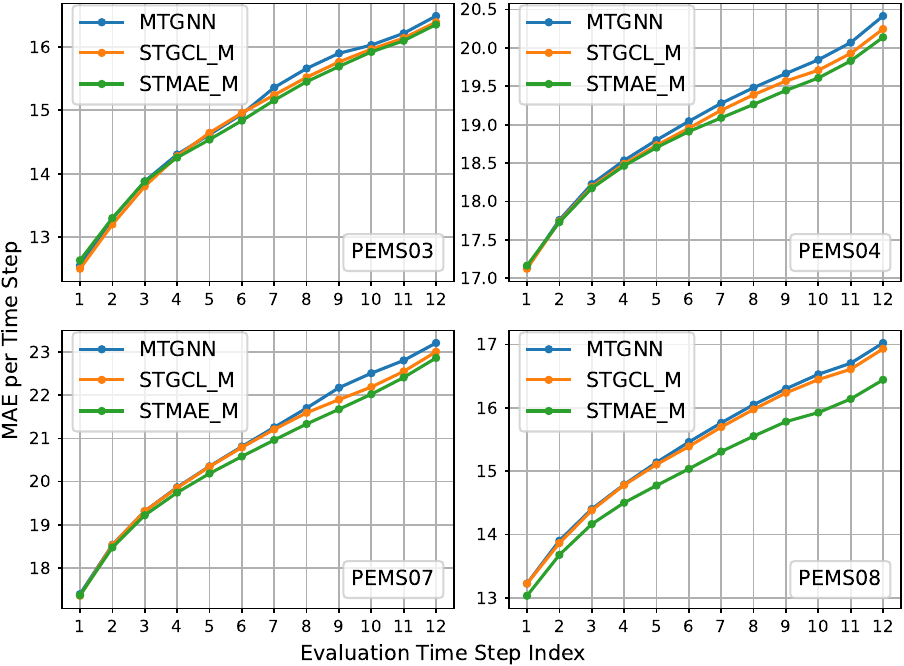}
\caption{Per-step MAE results of $\textrm{\name}_{\textrm{M}}$ compared with its corresponding base model MTGNN and $\textrm{STGCL}_{\textrm{M}}$.}
\label{fig::per_step_m}
\Description{}
\end{figure}
\begin{figure}[htbp]
\centering
\includegraphics[width=1.0\linewidth]{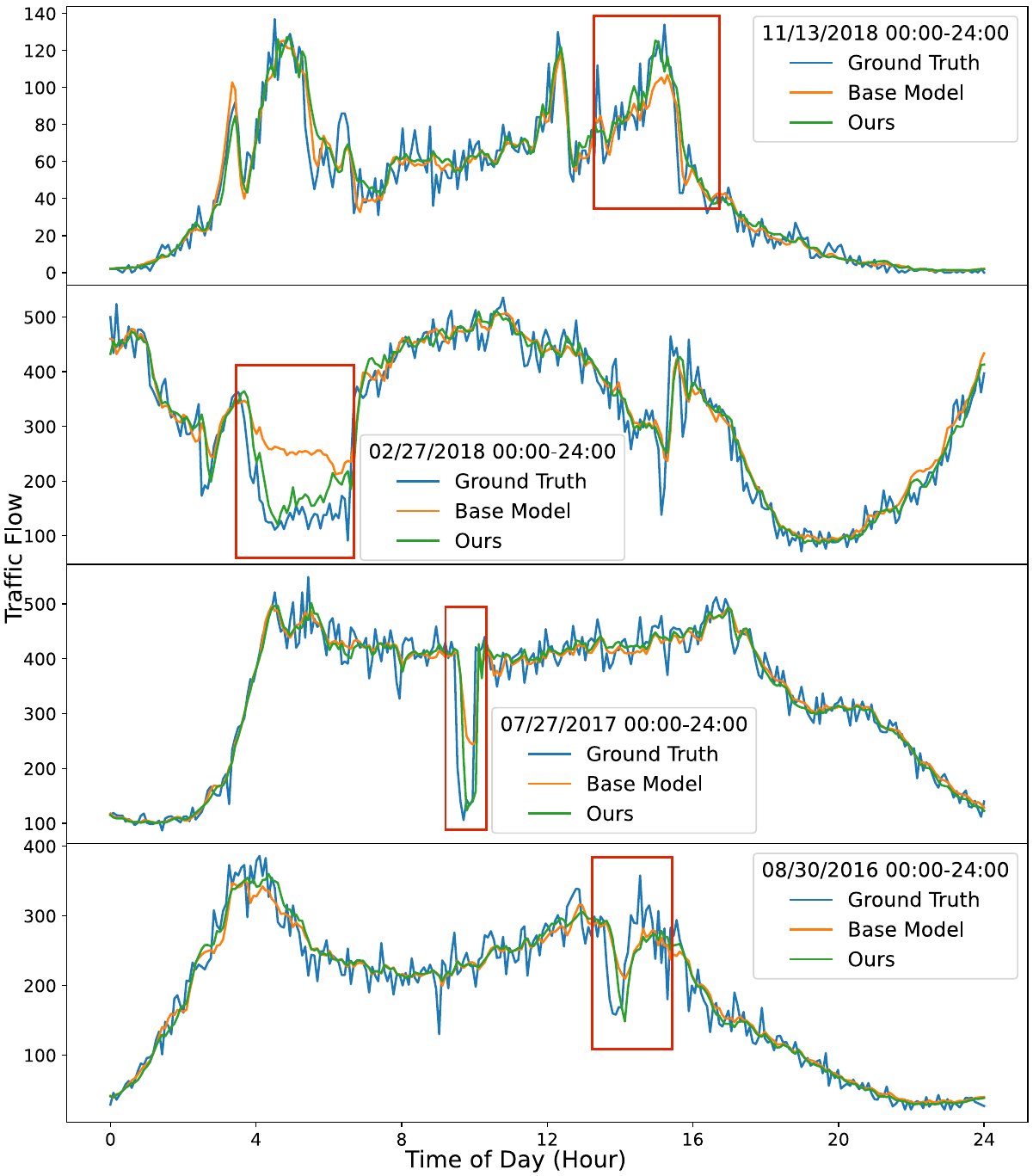}
\caption{Visualizations of $\textrm{\name}_{\textrm{A}}$ and AGCRN one-hour-ahead predictions on snapshots from PEMS03, PEMS04, PEMS07 and PEMS08 test sets.}
\label{fig::vis_a}
\Description{}
\end{figure}
\begin{figure}[!t]
\centering
\includegraphics[width=1.0\linewidth]{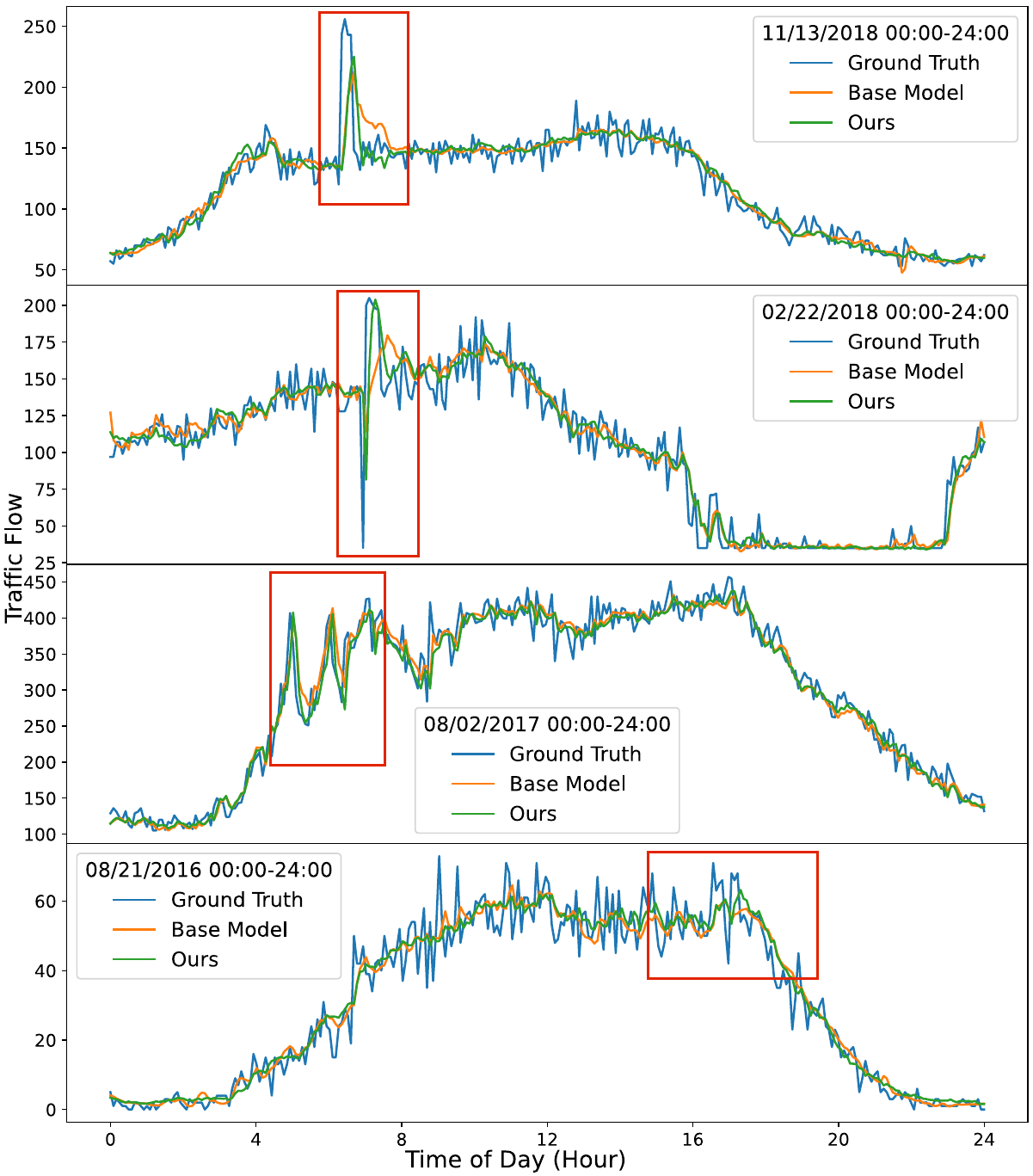}
\caption{Visualizations of $\textrm{\name}_{\textrm{D}}$ and DCRNN one-hour-ahead predictions on snapshots from PEMS03, PEMS04, PEMS07 and PEMS08 test sets.}
\label{fig::vis_d}
\Description{}
\end{figure}
In this section, we provide additional per-step MAE results, including (1) $\textrm{\name}_{\textrm{D}}$, DCRNN and $\textrm{STGCL}_{\textrm{D}}$ and (2) $\textrm{\name}_{\textrm{M}}$, MTGNN and $\textrm{STGCL}_{\textrm{M}}$ in \cref{fig::per_step_d,fig::per_step_m}. 
The additional results further confirms \name's ability in alleviating the performance degradation issue in spatial-temporal models with its generally enhanced performance over baselines.

\section{Additional Visualizations}
\label{appendix:vis}

\begin{figure}[htbp]
\centering
\includegraphics[width=1.0\linewidth]{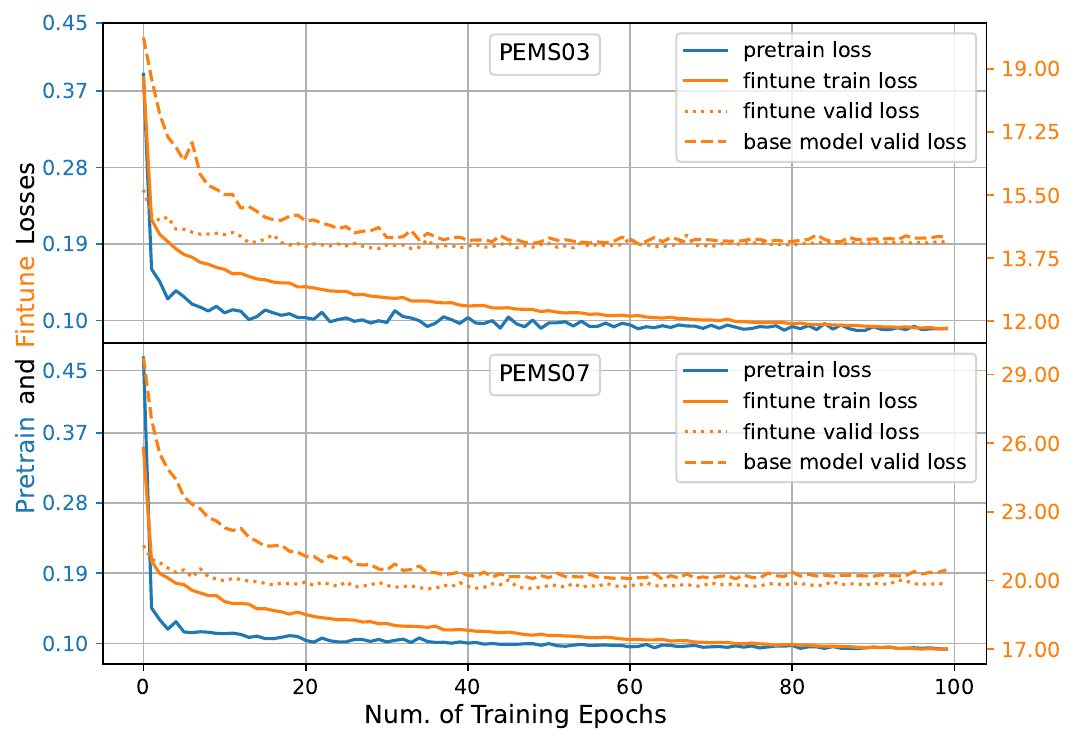}
\caption{Additional training and validation processes of $\textrm{\name}_{\textrm{A}}$ and AGCRN on PEMS03 and PEMS07 datasets.}
\label{fig::loss37}
\Description{}
\end{figure}

\begin{figure}[htbp]
\centering
\includegraphics[width=1.0\linewidth]{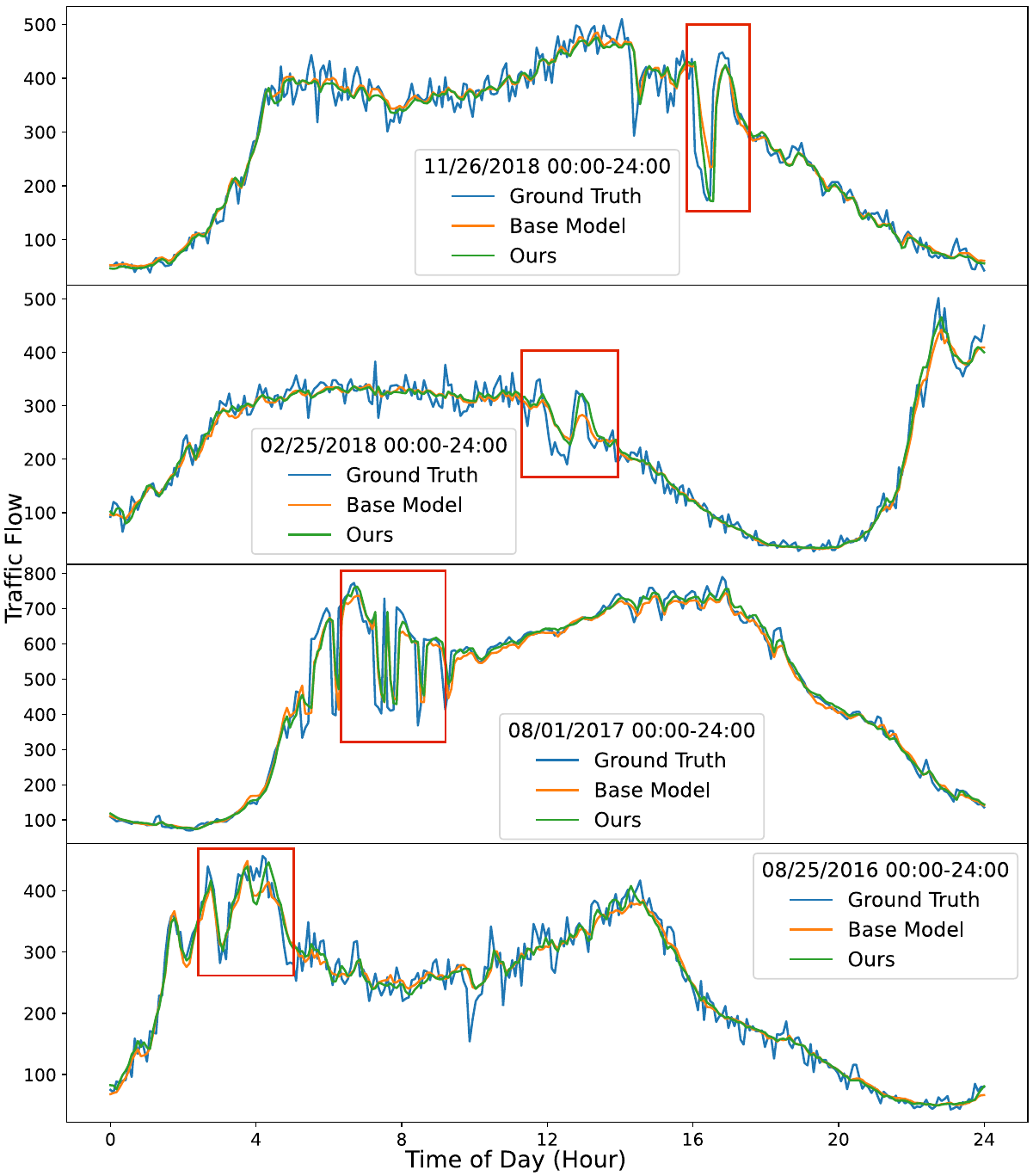}
\caption{Visualizations of $\textrm{\name}_{\textrm{M}}$ and MTGNN one-hour-ahead predictions on snapshots from PEMS03, PEMS04, PEMS07 and PEMS08 test sets.}
\label{fig::vis_m}
\Description{}
\end{figure}

In this section, we provide additional qualitative visualizations in \cref{fig::vis_a,fig::vis_d,fig::vis_m}. 
To further demonstrate \name's stability against data irregularities, we randomly select a subset of sensors and days from the PEMS03, PEMS04, PEMS07 and PEMS08 datasets, characterized by sudden traffic flow changes. 
We visualize one-hour-ahead predictions of \name~against the corresponding base models. 
From the areas highlighted by red rectangles, indicative of sudden flow changes, we can confirm that \name~is more robust to data irregularities, consistently outperforming the counterpart base models.

\section{Additional Ablation Results}
\label{appendix:ablation}

In this section, we include additional ablation results for evaluating \name’s overall design.
Experimental results are presented in \cref{tab:full_ablation}, with AGCRN serving as the backbone, while PEMS03 and PEMS07 are used as additional evaluation datasets. 
The results further underscores the advantages of our approach, where the advanced dual-masking strategy can create challenging pretext tasks for spatial-temporal models, thus improving their forecasting capacities. 
\begin{table}[t]\small
\centering
\caption{Additional ablation results of $\textrm{STMAE}_{\textrm{A}}$ on PEMS03 and PEMS07. Best results are \textbf{bolded} and second bests are \underline{underlined}.} 
\resizebox{\linewidth}{!}{
    \begin{tabular}{ c | ccc | ccc } 
      \hline
       & \multicolumn{3}{c}{PEMS03} & \multicolumn{3}{c}{PEMS07} \\
      \hline
      Variant & MAE & MAPE & RMSE & MAE & MAPE & RMSE \\
      \hline
      AGCRN & 15.47 & 15.26 & 27.06 & 20.64 & 8.80 & \underline{34.19} \\
      $\textrm{\name}_{\textrm{NT}}$ & 15.30 & 15.26 & \underline{26.84} & 20.52 & 8.81 & 34.28 \\
      $\textrm{\name}_{\textrm{NS}}$ & 15.27 & 15.30 & {27.05} & 20.42 & 8.63 & 34.29 \\
      $\textrm{\name}_{\textrm{U}}$ & \underline{15.25} & \underline{15.12} & 27.31 & \underline{20.38} & \underline{8.50} & 34.22 \\
      \hline 
      \rowcolor{gray}
      $\textrm{\name}_{\textrm{A}}$ & \textbf{15.09} & \textbf{{14.72}} & \textbf{{26.61}} & \textbf{{20.13}} & \textbf{{8.53}} & \textbf{{33.79}}  \\
      \hline
    \end{tabular}}
\label{tab:full_ablation}
\end{table}

\section{Additional Stability Study Results}
\label{appendix:stability}

\begin{figure}[htbp]
\centering
\includegraphics[width=1.0\linewidth]{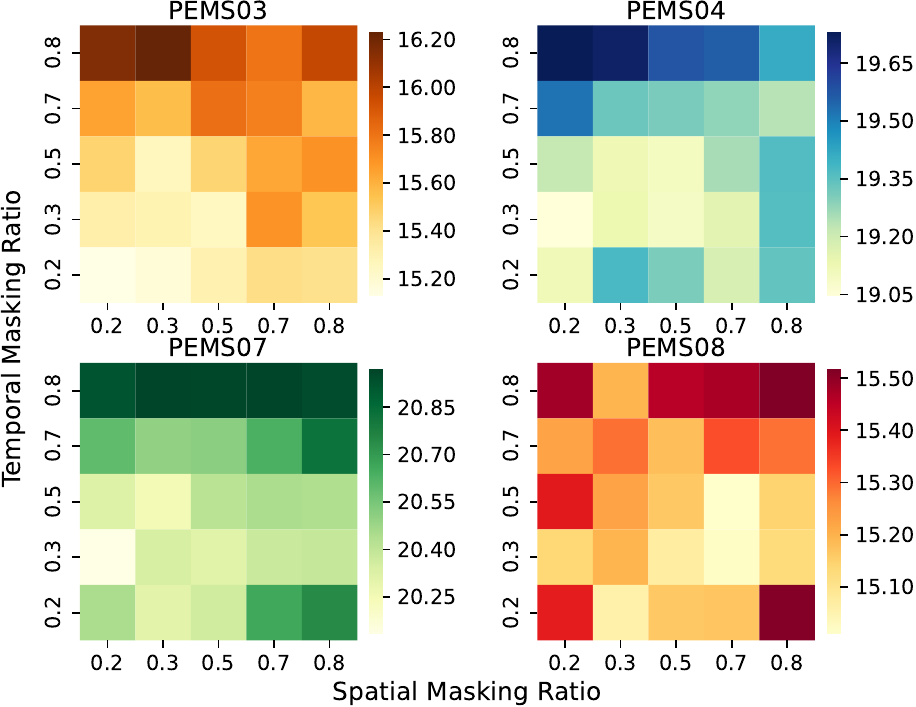}
\caption{Masking ratio sensitivity analysis of $\textrm{\name}_{\textrm{A}}$ on PEMS03, PEMS04, PEMS07 and PEMS08. The values denote the MAE metric. Lighter is better.}
\label{fig::mask}
\Description{}
\end{figure}

We provide additional experimental results for \name~stability study in \cref{fig::loss37}, including visualizations of the learning curves for both the pretraining and fine-tuning stages of $\textrm{\name}_{\textrm{A}}$ on PEMS03 and PEMS07, alongside the learning curve for the AGCRN backbone.

\clearpage

\section{Additional Masking Ratio Results}
\label{appendix:mask}

In this section, we include additional results for studying \name’s sensitivity to the spatial and temporal masking ratios.
As mentioned in the main paper, we explore the impact of varying these ratios from 20\% to 80\% when coupling STMAE with AGCRN, and experimental results for all four datasets: PEMS03, PEMS04, PEMS07 and PEMS08 are shown in \cref{fig::mask}, where the values indicate test set MAE performance.
From the heatmaps, we draw the same conclusion that \name~exhibits optimal performance when the temporal masking ratio is around 20\% to 30\% for all datasets. 
On the other hand, from the spatial perspective, STMAE performs the best with a spatial masking ratio of 70\% for PEMS08, while for other datasets, 20\% to 30\% is a more favorable choice. 
We attribute this difference to the greater physical structural density of PEMS08 as shown in \cref{tab:dataset_stat}.

\end{document}